\documentclass[10pt,twocolumn,letterpaper]{article}

\usepackage[pagenumbers]{iccv} 
\usepackage{circledsteps} 
\usepackage{expl3}
\usepackage{pgfplots} 
\usepackage{makecell} 
\usepackage{pifont}
\newcommand{\name}[0]{SegFit}

\usepackage[skip=5px]{caption}
\renewcommand{\paragraph}[1]{\vspace{5px}\noindent\textbf{#1}\,}

\definecolor{iccvblue}{rgb}{0.21,0.49,0.74}
\usepackage[pagebackref,breaklinks,colorlinks,allcolors=iccvblue]{hyperref}
\usepackage{multirow}
\usepackage{colortbl}
\usepackage{overpic}
\newcommand{\xmark}{\ding{55}}
\newcommand{\cmark}{\ding{51}}%

\usepackage{pgfplots}
\pgfplotsset{compat=1.17}
\usepackage{pgfplotstable}

\def\paperID{4574} 
\def\confName{ICCV}
\def\confYear{2025}

\title{Robust Human Registration with Body Part Segmentation on Noisy Point Clouds}

\author{
Kai Lascheit$^{1}$ \hspace{10px}
Daniel Barath$^{1,4}$ \hspace{10px}
Marc Pollefeys${^{1,2}}$ \hspace{10px}%
Leonidas Guibas${^{3}}$ \hspace{10px}
Francis Engelmann${^{3}}$
\vspace{5px}\\
\vspace{5px}
{\small
$^1$ETH Zurich
\hspace{15px}
$^2$Microsoft
\hspace{15px}
$^3$Stanford University
\hspace{15px}
$^4$Google
\hspace{15px}
}
}

\begin{document}

\twocolumn[{
\renewcommand\twocolumn[1][]{#1}%
\maketitle

\centering
\begin{small}
\includegraphics[width=\textwidth]{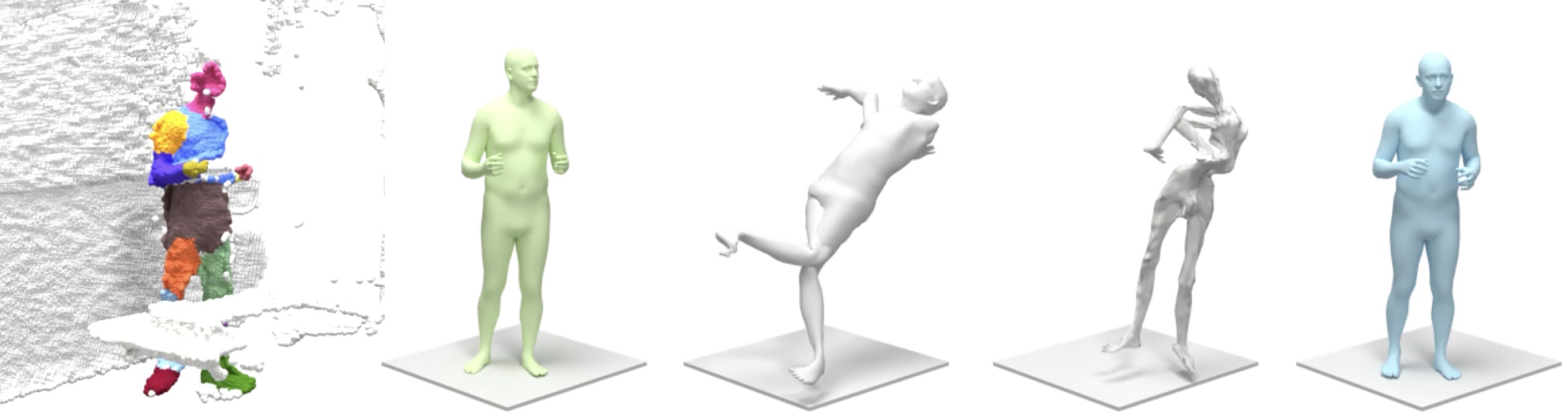}
\\
\begin{tabular}{cccccc}
\hspace{0.18\textwidth} & %
\hspace{0.18\textwidth} & %
\hspace{0.18\textwidth} & %
\hspace{0.18\textwidth} \\
{\textit{Input Point Cloud}}
& {\textit{Ground Truth}}
& {\textit{ArtEq}~\cite{arteq}}
& {\textit{NICP}~\cite{marin2025nicp}}
& {\textit{SegFit} (Ours)}
\end{tabular}
\end{small}
\captionof{figure}{
Our method SegFit reconstructs human poses from point clouds using body part segmentation and the SMPL-X model \cite{pavlakos2019expressive}.
We showcase SMPL-X fitting results on the EgoBody dataset \cite{zhang2022egobody}, and compare to the state-of-the-art methods ArtEq \cite{arteq} and NICP \cite{marin2025nicp}.
\vspace{25px}
}
\label{fig:teaser}}]

\begin{abstract}
\vspace{-20px}\\
Registering human meshes to 3D point clouds is essential for applications such as augmented reality and human-robot interaction but often yields imprecise results due to noise and background clutter in real-world data.
We introduce a hybrid approach that incorporates body-part segmentation into the mesh fitting process, enhancing both human pose estimation and segmentation accuracy.
Our method first assigns body part labels to individual points, which then guide a two-step SMPL-X fitting:
initial pose and orientation estimation using body part centroids,
followed by global refinement of the point cloud alignment.
Additionally, we demonstrate that the fitted human mesh can refine body part labels,
leading to improved segmentation.
Evaluations on the cluttered and noisy real-world datasets InterCap, EgoBody, and BEHAVE show that our approach significantly outperforms prior methods in both pose estimation and segmentation accuracy.
Code and results are available on our project website: \href{https://segfit.github.io}{https://segfit.github.io}
\end{abstract}    
\vspace{-10px}
\section{Introduction}
Registering parametric human meshes to 3D point clouds requires estimating both human pose and shape parameters. Accurately capturing the nuances of human movement and form is essential for various applications, from generating realistic virtual avatars~\citep{thalmann2012crowd,slater2016enhancing} to enhancing human-robot interactions~\citep{argall2009survey, lepert2025phantom}. While fitting a parametric human body model to 3D data (e.g., from LiDAR or Kinect) is a crucial step in these applications, the absence of contextual cues and the presence of noisy, cluttered environments often result in suboptimal fits.
In contrast, this work explores how leveraging body parts can significantly improve fitting accuracy. Towards that end, we introduce a novel method that integrates body part segmentation with pose fitting for partial, noisy, and cluttered 3D point clouds.

In recent years, parametric models like SMPL~\cite{SMPL2015} and its extension SMPL-X~\citep{pavlakos2019expressive} have become the de facto standard for describing 3D human poses and shapes.
These models are typically fitted to point cloud data using either gradient-based optimization~\citep{single-rgbd-fitting, bogo2016keep, pavlakos2019expressive} or neural networks~\citep{kolotouros2019learning, Kocabas2020VIBE}.
Advancements in body part segmentation—assigning anatomical labels to points in a 3D cloud—have unlocked new opportunities in computer graphics, healthcare, and autonomous systems. However, existing methods often struggle in real-world scenarios involving complex poses, occlusions, multi-person interactions, and human-object occlusions.

Many approaches rely on synthetic datasets~\citep{mahmood2019amass, andriluka20142d}, leading to performance degradation when faced with the variability of in-the-wild data~\citep{mehta2017vnect}.
Furthermore, these methods frequently fail to generalize to unseen body poses and varying sensor noise, limiting their effectiveness in practical applications.
To address these limitations, we propose a hybrid framework that merges pose fitting and body part segmentation, allowing each to refine the other. This refinement not only improves pose fitting but also enhances the initial segmentation, enabling further fine-tuning of the segmentation network.
We begin with an initial segmentation from the Human3D network~\citep{Takmaz_2023_ICCV}, which provides a coarse assignment of points to body parts and is already fine-tuned on in-the-wild data ~\citep{zhang2022egobody}.
This segmentation serves as the foundation for a two-step optimization procedure that fits the SMPL-X model to the human point clouds.
First, body-part centroids guide an approximate alignment of the model pose and orientation~\citep{bogo2016keep, xiang2019monocular}, establishing a robust initial configuration.
Second, the model is refined by considering all points in the cloud, thereby capturing more nuanced pose and shape information~\citep{pavlakos2019expressive, kolotouros2019learning}.
Throughout this process, we incorporate a pose prior~\citep{pavlakos2019expressive} to ensure anatomically plausible body configurations, mitigating errors caused by occlusions or missing data.
After fitting, we reassign body parts to the 3D point cloud via majority voting over nearest neighbors~\citep{ge20193d, Remelli2020MeshSDF}, yielding a segmentation more accurate than the initial prediction. By uniting part segmentation and mesh fitting, our approach more effectively generalizes to diverse environments and remains robust under challenging conditions.

We evaluate our approach on three complex datasets -- InterCap~\citep{huang2024intercap}, EgoBody~\citep{zhang2022egobody}, and BEHAVE~\citep{bhatnagar22behave} -- featuring occlusions, multiple interacting subjects, and human-object interactions.
Compared to other leading methods~\citep{arteq, marin2025nicp, Takmaz_2023_ICCV}, we observe an up to \textit{tenfold} boost in pose modeling accuracy and an up to \textit{22\%} gain in segmentation accuracy.

Our contributions are summarized as follows:
\begin{itemize}[noitemsep,leftmargin=1.2em]
    \item \textbf{Segmentation-Based Pose Fitting.} A unified approach that integrates human pose fitting with body part segmentation on point clouds, increasing the accuracy and robustness of the fitting process in the presence of noisy point clouds.
    {\item \textbf{Pose Fitting-Enhanced Segmentation.} We utilize fitted SMPL-X meshes to refine body-part labels, leading to more precise segmentation. Additionally, we demonstrate how these improved segmentations can be leveraged for self-supervised fine-tuning of a segmentation network.}
\end{itemize}

Our work advances the state-of-the-art in 3D human body fitting and segmentation for point clouds, promising more faithful human representations in real-world, unstructured environments.

\section{Related Work}

Estimating human body pose and shape from 3D point clouds is vital in computer vision, with applications in virtual reality, animation, and human-computer interaction. While extensive research has been conducted on fitting parametric human models to 2D images~\cite{bogo2016keep, kanazawa2018end, kolotouros2019learning}, we focus on methods that directly operate on 3D point cloud data. Point clouds capture detailed geometric information and avoid the ambiguities inherent in 2D projections, making them valuable for precise human modeling. \\

Several approaches have been developed to fit human poses to point clouds. \citet{bhatnagar2020combining} introduced \textit{IP-Net}, which combines implicit representations with parametric models to reconstruct clothed human bodies from partial scans. IP-Net learns a continuous occupancy field representing the human body, allowing for detailed reconstructions even with incomplete data. 
\citet{PTF} proposed \textit{PTF}, a method that fits SMPL models to point clouds by considering local geometric features. By leveraging local point distributions, PTF improves fitting accuracy in areas with high curvature or fine-grained details.
\citet{zuo2021self} presented a self-supervised approach for 3D human motion reconstruction from depth sequences. Their method leverages temporal coherence without requiring ground truth annotations, effectively reconstructing dynamic human motions.
\citet{cai2023pointhps} developed \textit{PointHPS}, a hierarchical point-based network that directly regresses SMPL parameters from point clouds. PointHPS achieves state-of-the-art results using a point-based encoder-decoder architecture.
While these methods can effectively reconstruct common poses, they struggle to accurately capture complex or non-standard poses, particularly those involving occlusions or extreme body configurations. \\

Recently, \citet{marin2025nicp} introduced \textit{NICP (Neural ICP)}, which bridges classic ICP with learnable shape representations. NICP iteratively refines a neural deformation field at inference time to better align human models with input point clouds, demonstrating notable improvements in registration accuracy across challenging poses and noisy scans.
Moreover, \citet{arteq} proposed \textit{ArtEq}, a part-based SE(3)-equivariant neural network for SMPL fitting. Unlike previous learning-based methods that struggle with out-of-distribution poses, ArtEq explicitly incorporates SE(3) invariance and equivariance to improve generalization. The method achieves state-of-the-art accuracy on the PosePrior subset of AMASS \cite{mahmood2019amass} and is significantly faster than previous methods. \\

Despite these advances, common challenges persist. Methods often rely on large annotated datasets or specialized training for generalization, which may be challenging to scale. Regression-based approaches can miss subtle body shape variations or fail under significant occlusions. Optimization-based algorithms are sensitive to initialization and can get stuck in local minima, especially for complex poses with self-contact or multiple people interacting. Our work addresses these challenges by integrating body part segmentation into the fitting process, leveraging semantic cues to distinguish between symmetric limbs and reduce orientation ambiguities. The segmentation-informed initialization obviates the need for multiple optimization runs (as in SMPLify-X~\cite{pavlakos2019expressive}), ensuring more robust convergence without exhaustive restarts.

\begin{figure}[t]
    \begin{center}
    \includegraphics[width=0.49\textwidth,trim={10px 10px 10px 10px},clip]{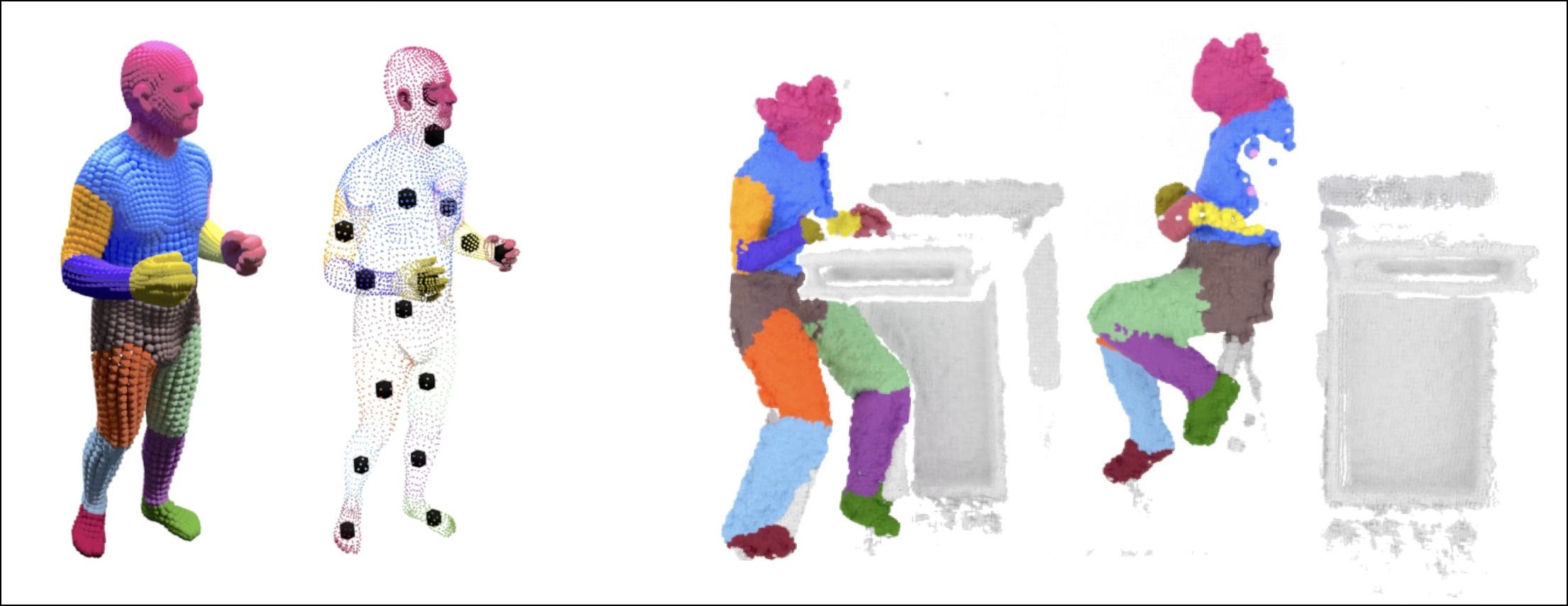}
    \end{center}
    \begin{small}\textit{\hspace{5px}Body-Parts \hspace{3px} Centroids \hspace{25px} Segmentation of two humans}\end{small}\\
    \vspace{-8px}
    \caption{The 15 body parts and their centroids \emph{(left)}. Example body parts segmentation of two humans from our \name{} \emph{(right).}}
    \label{fig:centroids}
\end{figure}

\section{Method}

We propose a framework for fitting the SMPL-X parametric body model to real-world 3D point clouds by leveraging initial body part segmentation.
Our method is built around two key processes:
a robust model initialization guided by body part centroids and a subsequent iterative fitting phase that refines pose and shape parameters.
We also show that we can use the fitted body mesh to refine the initial body-part segmentation based on a nearest-neighbor approach.

\subsection{Problem Definition}

Given a 3D point cloud $\mathcal{P} = \{\mathbf{p}_i\}_{i=1}^N$ representing a cluttered scene containing one or more human subjects, our goal is to estimate the pose and shape parameters of the SMPL-X model~\citep{pavlakos2019expressive} to ensure accurate alignment between the resulting 3D meshes and the human point clouds.
A key challenge lies in identifying which points in the cluttered scene correspond to the human body surface, compounded by the inherent symmetry of the human body—leading standard optimization methods to converge to suboptimal solutions without proper initialization. To tackle these issues, our approach leverages human body-part segmentation from Human3D~\citep{Takmaz_2023_ICCV}, providing essential semantic guidance for both initialization and optimization.

\subsection{Overview of the Approach}

Our approach consist of four major steps:
\begin{enumerate}[noitemsep,leftmargin=1.5em]
    \item \textbf{Initial Body Part Segmentation:}
    Utilize Human3D~\citep{Takmaz_2023_ICCV} to assign a body part index to each point in $\mathcal{P}$, yielding a coarse but informative segmentation of the point cloud.
    \item \textbf{Model Initialization:} Calculate the body part centroids (see Figure \ref{fig:centroids}) from the segmentation and align them with the corresponding centroids of an SMPL-X template to obtain a stable initial pose and orientation.
    \item \textbf{Model Fitting:} Refine the SMPL-X parameters by optimizing a multi-term objective that balances data fidelity and pose/shape regularization over the entire human point cloud.
    \item \textbf{Enhanced Body-Part Segmentation:} Label body parts using nearest-neighbor majority voting on the fitted mesh, refining the body-part segmentation from the initial network output.
\end{enumerate}
Below, we describe each component in detail.

\subsection{Initial Body Part Segmentation}

We begin by segmenting the input point cloud $\mathcal{P}$ with Human3D~\citep{Takmaz_2023_ICCV}, a state-of-the-art network that predicts a body part label $s_i \in \{1,\dots,K\}$ for every point $\mathbf{p}_i \in \mathcal{P}$. This step encodes high-level semantic cues about the spatial organization of the human body in 3D, enabling subsequent stages to distinguish between symmetric limbs and reduce ambiguity in pose initialization.

\subsection{Model Initialization}
\label{sec:model_init}
A common pitfall in human model fitting is suboptimal initialization, which can hinder or derail convergence. Rather than performing multiple fitting trials with varied initial orientations, we leverage the body part segmentation to establish a direct, data-driven initialization. Specifically:
\begin{enumerate}[leftmargin=1.5em]
    \item Compute centroid $\mathbf{c}_k^{\text{scan}} = \frac{1}{N_k}\sum_{i=1}^{N_k} \mathbf{p}_i$ for each body part $k$, where $N_k$ is the number of points labeled $k$.
    \item Identify the corresponding centroids in the SMPL-X template $\mathcal{M}_0$, denoted by $\mathbf{c}_k^{\text{model}}$.
    \item Compute a global rotation $\mathbf{R}_0$ and translation $\mathbf{t}_0$ that align $\mathbf{c}_k^{\text{model}}$ to $\mathbf{c}_k^{\text{scan}}$, providing a well-informed initial pose for the subsequent optimization.
\end{enumerate}
This centroid-based matching approach effectively addresses orientation ambiguities by exploiting structural cues in the data, similar to finding the corner pieces of a puzzle before refining the interior.

\subsection{Model Fitting}
\label{sec:model_fitting}

\paragraph{SMPL-X Parameterization.}
The SMPL-X model~\citep{pavlakos2019expressive} is parameterized by pose $\boldsymbol{\theta}\!\in\!\mathbb{R}^{\!3J}$, shape $\boldsymbol{\beta}\!\in\!\mathbb{R}^{B}$, and global translation $\mathbf{t}\!\in\!\mathbb{R}^3$, where $J$ is the number of joints and $B$ is the dimension of the shape space. To avoid implausible configurations, we employ VPoser~\citep{pavlakos2019expressive}, a learned human pose prior that maps $\boldsymbol{\theta}$ to a latent space with higher-level constraints on body articulation.

\paragraph{VPoser Prior.}
VPoser~\citep{pavlakos2019expressive} is a variational autoencoder (VAE)-based prior for human body pose which has been trained on the AMASS~\citep{mahmood2019amass} dataset, capturing a large range of natural human poses. It applies an encoder-decoder architecture to transform SMPL-X parameters into a compact latent space. We optimize the SMPL-X parameters in this latent space to constrain our method to realistic human poses, as is common in optimization approaches to human mesh fitting~\citep{pavlakos2019expressive, PROX:2019}.

\paragraph{Objective Function.}
We refine $\boldsymbol{\theta}$, $\boldsymbol{\beta}$, and $\mathbf{t}$ by minimizing a combined energy:
\begin{equation}
\label{eq:obj}
\mathcal{L} = \lambda_{\mathrm{data}} \mathcal{L}_{\mathrm{data}} + \lambda_{\mathrm{pose}} \mathcal{L}_{\mathrm{pose}} + \lambda_{\mathrm{shape}} \mathcal{L}_{\mathrm{shape}},
\end{equation}
with hyperparameters $\lambda_{\mathrm{data}}$=$1$, $\lambda_{\mathrm{pose}}$=$0.5$, $\lambda_{\mathrm{shape}}$=$0.5$.
A parameter study is presented in the experiments (Sec.~\ref{sec:experiments_analysis}). Below, we provide a description for each term.

\textit{Data Term} 
quantifies alignment between the model surface and $\mathcal{P}$. We adopt a robust, one-sided Chamfer distance with a Huber loss to make it robust as follows:
\begin{equation}
    \label{eq:data}
    \mathcal{L}_{\mathrm{data}} = \sum_{k=1}^K \sum_{i=1}^{N_k} \min_{\mathbf{v}\in \mathcal{V}_k} \mathcal{L}_{\mathrm{Huber}}(\mathbf{p}_i - \mathbf{v}),
\end{equation}
where $\mathcal{V}_k$ denotes vertices belonging to body part $k$. A one-sided formulation prioritizes model-to-data consistency and mitigates erroneous penalization of regions without corresponding sensor capture.

\textit{Pose Term} 
regularizes poses around the VPoser prior, encouraging kinematically realistic articulation:
\begin{equation}
\mathcal{L}_{\mathrm{pose}} = \|\boldsymbol{\theta} - \boldsymbol{\theta}_0\|_2^2,
\end{equation}
where $\boldsymbol{\theta}_0$ is the default pose of the SMPL-X model.
    
\textit{Shape Term} 
constrains shape coefficients to reasonable magnitudes to describe a realistic human shape as follows:
\begin{equation}
    \mathcal{L}_{\mathrm{shape}} = \|\boldsymbol{\beta}\|_2^2.
\end{equation}

We employ the Adam optimizer with early stopping (200 maximum steps) to minimize Eq.~\eqref{eq:obj}. This yields a refined pose and shape that closely aligns the SMPL-X body model to the input data.

\subsection{Enhanced Part Segmentation}

After fitting, we reassign part labels to each point $\mathbf{p}_i$ using majority voting among its nearest-neighbor vertices on the SMPL-X mesh. Formally, let $\{\mathbf{v}_1, \ldots, \mathbf{v}_n\}$ be the $n$ closest mesh vertices to $\mathbf{p}_i$, each labeled by a body part index. We obtain the new label $s_i$ via:
\begin{equation}
    s_i = \arg\max_{k \in \{1,\dots,K\}} \sum_{j=1}^{n} \alpha_j \,\delta\bigl(\text{label}(\mathbf{v}_j) = k\bigr),
\end{equation}
where $\alpha_j$ is an inverse distance weight, and $\delta(\cdot)$ is the Kronecker delta. This label reassignment leverages the accurate surface alignment from the fitted model, producing a more reliable body part segmentation than the initial one.

\subsection{Summary}

By integrating pose fitting with segmentation, our method achieves robust initializations and more precise model alignments.
This synergy surpasses approaches that treat pose estimation and segmentation separately,
particularly in real-world scenarios where occlusions, noise, and high variability challenge purely data-driven or optimization-based techniques.
\section{Experiments}

In this section, we first compare our approach \name{} with state-of-the-art human registration methods on three challenging real-world datasets (Sec.~\ref{sec:experiments_sota}).
We then show how \name{} also improves 3D human body-part segmentation (Sec.~\ref{sec:experiments_segmentation}).
Next, we provide detailed analysis experiments to understand the importance of human body part segments for human pose estimation, and the effects of varying optimization strategies (Sec.~\ref{sec:experiments_analysis}). Finally, we show qualitative results of \name{} on \emph{in-the-wild} environments (Sec.~\ref{sec:experiments_qualitative}).
    
\subsection{Comparing with State-of-the-Art Methods}
\label{sec:experiments_sota}

\begin{figure}[!b]
\centering
\begin{overpic}[abs,unit=1mm,width=\linewidth,trim=0 1.5 1.5 1.5, clip]{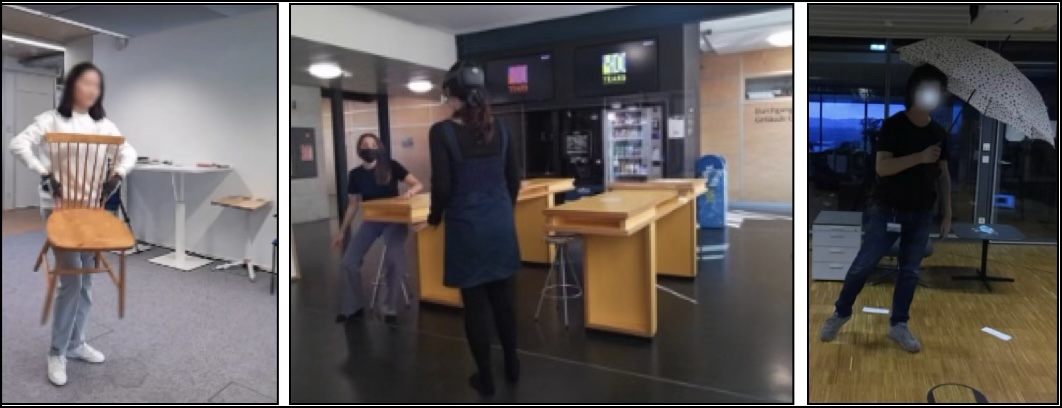}
\put(5,2){\small \color{white}BEHAVE}
\put(39,2){\small \color{white}EgoBody}
\put(67,2){\small \color{white}InterCap}
\end{overpic}
\caption{\textbf{Datasets.} Example scenes from BEHAVE~\cite{bhatnagar22behave}, EgoBody~\cite{zhang2022egobody}, and InterCap~\cite{huang2024intercap}. We show RGB images for illustration only, all experiments are performed on single-view depth maps.}
\label{fig:datasets}
\end{figure}

\paragraph{Datasets.}
To evaluate the robustness of our approach, we rely on three challenging ``in-the-wild" datasets \cite{bhatnagar22behave, zhang2022egobody, huang2024intercap}, which feature cluttered scenes, occlusions, and noisy, partial observations (see Fig.~\ref{fig:datasets}). These datasets provide a more realistic setting compared to controlled ``in-the-lab" datasets \cite{mahmood2019amass, yin2023hi4d, dfaust}, where humans are isolated from the background and captured with high-quality cameras from multiple-view points in empty scenes.

\emph{EgoBody}~\cite{zhang2022egobody} is a large-scale dataset capturing ground-truth 3D human motions during social interactions in natural 3D environments. It includes up to two individuals interacting with each other and their surroundings, resulting in significant occlusions, partial observations, and challenges in disentangling humans from the background.
\emph{BEHAVE}~\cite{bhatnagar22behave} is a full-body human-object interaction dataset containing multi-view RGB-D sequences and annotated human meshes in natural environments.
It features a single person interacting closely with various objects, leading to challenging poses and strong occlusions.
\emph{InterCap}~\citep{huang2024intercap} is another large-scale dataset focusing on human-object interactions, similar to BEHAVE.
Each scene involves a single human interacting with one out of ten different object types.
All three datasets are captured with multi-view Kinect RGB-D sensors which simplifies the accurate ground truth human mesh annotation. However, our experiments rely solely on single-view depth frames -- more representative of real-world applications -- leading to significant occlusions from objects in the scene and partial human observations.

\begin{figure}[t]
\centering
\small
\includegraphics[width=\linewidth]{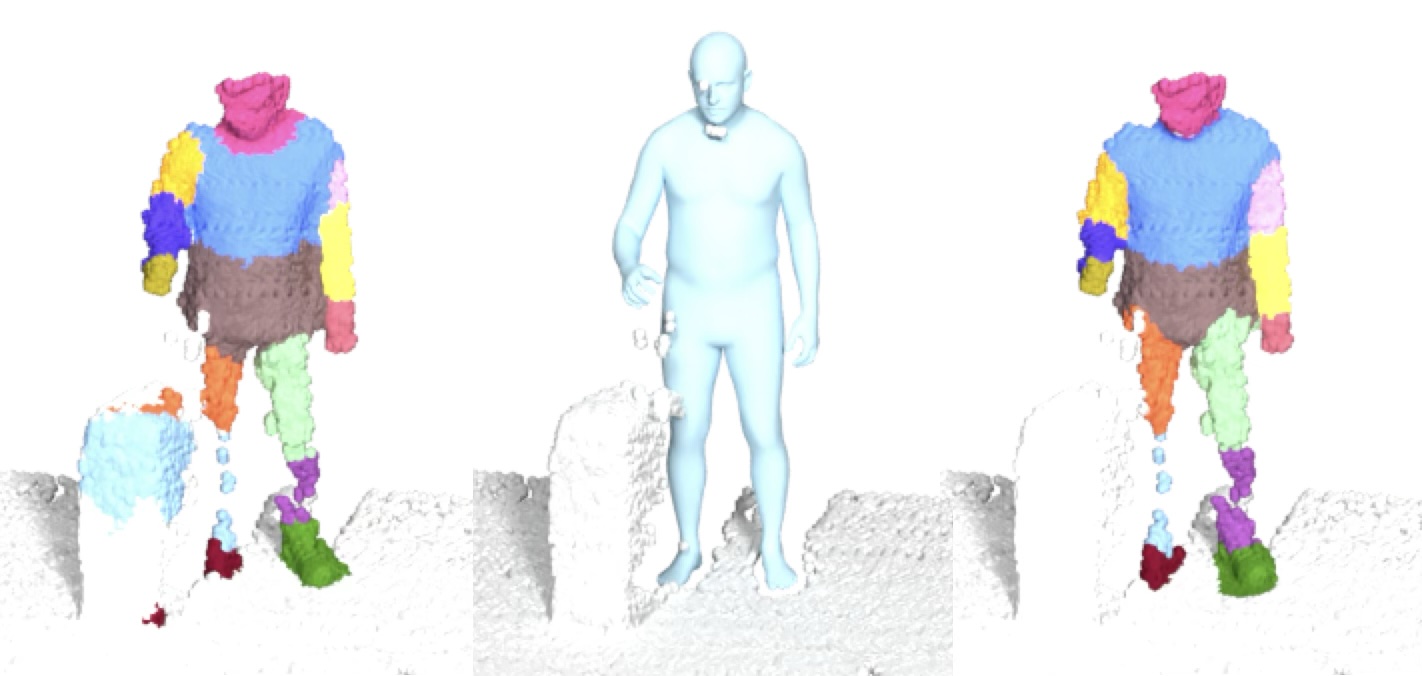}    
\vspace{15px}
\begin{tabular}{ccc}
 \textit{Human3D}~\cite{Takmaz_2023_ICCV} & \textit{SegFit Mesh (Ours)} & \textit{SegFit (Ours)}\\
 \hspace{2.3cm} & \hspace{2.3cm} & \hspace{2.3cm}\\
\end{tabular}
\vspace{-15px}
\caption{\textbf{Refined Body-Part Segmentation.}
Example output of our \name{} on the InterCap~\cite{huang2024intercap} dataset.
We show the initial body-part segmentation from Human3D~\cite{Takmaz_2023_ICCV} \emph{(left)}, our registered human mesh \emph{(center)}, and the improvied human body-part segmentation based on nearest neighbors majority voting \emph{(right)}. Notice that, at first, the suitecase is mistakenly labeled as the left leg. However, despite the occlusion, our approach successfully corrects the human pose and refines the body-part segmentation.}
\label{fig:improved_segmentation}
\end{figure}

\begin{table*}[t!]
\vspace{5px}
\small
\setlength{\tabcolsep}{9.5pt}
\centering
\begin{tabular}{l ccc c ccc c ccc}
\toprule
 &
\multicolumn{3}{c}{\textbf{BEHAVE}~\cite{bhatnagar22behave}} &&
\multicolumn{3}{c}{\textbf{EgoBody}~\cite{zhang2022egobody}} &&
\multicolumn{3}{c}{\textbf{InterCap}~\cite{huang2024intercap}}\\
\cmidrule{2-4}  \cmidrule{6-8} \cmidrule{10-12}
 &
V2V & MPJPE & Time &&
V2V & MPJPE & Time &&
V2V & MPJPE & Time \\
Method & {\small in mm}& {\small in mm}& {\small in s}&&
{\small in mm}& {\small in mm}& {\small in s}&&
{\small in mm}& {\small in mm}& {\small in s}\\
\midrule
ArtEq~\cite{arteq} & $140.6$ & $162.4$ & $\mathbf{0.105}$ && $538.7$ & $605.7$ & $\mathbf{0.098}$ && $422.0$ & $515.0$ & $\mathbf{0.103}$ \\
NICP~\cite{marin2025nicp} & $59.9$ & -- & $33.14$ && $232.1$ & -- & $17.2$ && $257.7$ & -- & $22.2$\\
SegFit (Ours) & $\mathbf{37.0}$ & $\mathbf{30.7}$ & $1.86$ && $\mathbf{47.9}$ & $\mathbf{42.2}$ & $1.79$ && $\mathbf{147.2}$ & $\mathbf{140.7}$ & $1.37$\\
\bottomrule
\end{tabular}
\caption{Pose and shape scores of \name{} in comparison to NICP~\cite{marin2025nicp} and ArtEq~\cite{arteq}
on the BEHAVE~\cite{bhatnagar22behave}, EgoBody~\cite{zhang2022egobody}, and InterCap~\cite{huang2024intercap} datasets.
Metrics are vertex-to-vertex (V2V) distance, mean-per-joint-position-error (MPJPE) and average runtime per-human.}
\label{tab:accuracy_runtime}
\end{table*}

\paragraph{Baseline Methods.}
We compare our \name{} with the most recent state-of-the-art methods for registering SMPL human meshes to 3D point clouds:
\emph{ArtEq}~\cite{arteq} introduces articulated SE(3)-equivariance for SMPL model fitting, enabling generalization to unseen poses by learning part-based transformations instead of global ones.
It combines SO(3)-invariant part detection with pose- and shape-equivariant regression, leveraging self-attention layers to preserve equivariance. 
\emph{NICP}~\cite{marin2025nicp} is a ICP-style self-supervised approach tailored to neural fields, enabling robust and scalable 3D human registration across diverse shapes and datasets without requiring manual annotations improving over comparable prior approaches~\cite{bhatnagar2020combining, bhatnagar2020loopreg, PTF}.

To ensure a fair comparison, we apply one adjustment:
since ArtEq and NICP are not designed for multi-human or cluttered scenes with backgrounds and occlusions,
we first isolate human instances using Human3D's human instance segmentation results before processing the resulting human point clouds with NICP and ArtEq. This ensures that all methods process only points that belong to humans as predicted by Human3D~\cite{Takmaz_2023_ICCV}.

\paragraph{Metrics.}
We follow prior work \cite{arteq, PTF} to evaluate the accuracy of the registered SMPL~\cite{SMPL2015} human model.
Shape error is measured as the Euclidean distance between corresponding mesh vertices, \ie{},
the vertex-to-vertex (V2V) error in mm.
Pose accuracy is evaluated as the mean per joint position error (MPJPE) in mm between the fitted and ground-truth SMPL models.
We also report the average processing time required to fit a single human instance.
For body-part segmentation, we follow \citep{Takmaz_2023_ICCV, arteq} and report accuracy (Acc), intersection over union (IoU), and mean average precision (mAP).

\label{sec:pose_estimation_results}

\paragraph{Results.}
Table~\ref{tab:accuracy_runtime} presents the results for human pose and shape fitting across all evaluated datasets. MPJPEs are not reported for NICP, as this method only predicts SMPL vertex positions. \name{} consistently outperforms all prior methods by a substantial margin, demonstrating superior generalization to a wide range of diverse and noisy real-world point clouds.

Overall, SegFit achieves the highest performance on BEHAVE, while its accuracy is lower on InterCap, which presents the greatest challenges due to its complex human poses and strong human-object occlusions. In comparison to NICP, SegFit shows notably improved performance on EgoBody, where severe occlusions -- such as instances where only a foot remains visible while the entire leg is hidden -- highlight the importance of body-part segmentation as a critical signal for more accurate human registration.
Lastly, while ArtEq is considerably faster due to its optimization-free nature, SegFit still offers a significant efficiency advantage over NICP, delivering at least a tenfold reduction in runtime while maintaining strong accuracy.

\begin{table}[t!]
\centering
\small
\vspace{10px}
\setlength{\tabcolsep}{5pt}
\begin{tabular}{ccccc}
\toprule
\textbf{Method} &
\textbf{Metric} &
\textbf{BEHAVE} &
\textbf{EgoBody} &
\textbf{Intercap} \\
\midrule
Human3D & \multirow{2}{*}{Acc} & $74.61$\% & $77.06$\% & $55.85$\% \\
 + SegFit & & $\mathbf{91.22}$\% & $\mathbf{84.71}$\% & $\mathbf{67.86\%}$ \\
\midrule
 Human3D & \multirow{2}{*}{IoU} & $57.38$\% & $62.89$\% & $45.37$\% \\
 + SegFit & & $\mathbf{77.26}$\% & $\mathbf{69.48}$\% & $\mathbf{55.13}$\% \\
\midrule
 Human3D & \multirow{2}{*}{mAP} & $73.42$\% & $73.98$\% & $59.54$\% \\
 + SegFit & & $\mathbf{85.74}$\% & $\mathbf{77.20}$\% & $\mathbf{67.69}$\% \\
\bottomrule
\end{tabular}
\caption{Scores of human body part segmentation before and after SegFit.
Metrics are accuracy (Acc), intersection over union (IoU), and mean average precision (mAP).}
\label{tab:body_part_segmentation_improvements}
\end{table}

\subsection{Refined Body-Part Segmentation}
\label{sec:experiments_segmentation}

Beyond human registration, we also assess \name{}'s enhanced body-part segmentation and compare it to the initial segmentation from Human3D~\cite{Takmaz_2023_ICCV}, as shown in Table~\ref{tab:body_part_segmentation_improvements}. After fitting the SMPL-X model, we refine the body-part labels by reassigning them through majority voting among the nearest model vertices, leading to improved segmentation quality.
Human3D is first pre-trained on synthetic data and then fine-tuned on Kinect RGB-D sensor data, the same sensor used across all datasets, resulting in a minimal generalization gap. The improved results obtained by SegFit show that model-based segmentation still leads to substantial improvements in segmentation accuracy, of approximately $17$\%, $8$\%, and $12$\% on the BEHAVE, EgoBody, and InterCap datasets, respectively. This demonstrates that our method effectively enhances body-part segmentation by leveraging the fitted models.
Figure~\ref{fig:improved_segmentation} provides an example where Human3D initially misclassified a suitcase as the right leg. After applying SegFit, it was correctly identified as part of the background, demonstrating the effectiveness of our approach in resolving segmentation errors.

\begin{table}[b]
\centering
\small
\begin{tabular}{l@{\hskip 8pt}ccccc@{\hskip 8pt}}
\toprule
\textbf{Human3D Model} & \textbf{Metric} & \textbf{BEHAVE} & \textbf{InterCap} \\
\midrule
w/o Fine-Tuning & \multirow{2}{*}{Accuracy} & 74.61\% & 55.85\% \\
w/\phantom{ } Fine-Tuning & & \textbf{89.96}\% & \textbf{67.98}\% \\
\midrule
w/o Fine-Tuning & \multirow{2}{*}{IoU} & 57.38\% & 45.37\% \\
w/\phantom{ } Fine-Tuning & & \textbf{75.36}\% & \textbf{56.71}\% \\
\midrule
w/o Fine-Tuning & \multirow{2}{*}{AP} & 73.42\% & 59.54\% \\
w/\phantom{ } Fine-Tuning & & \textbf{84.56}\% & \textbf{68.18}\% \\
\bottomrule
\end{tabular}
\caption{Segmentation performance of Human3D~\citep{Takmaz_2023_ICCV} before and after fine-tuning in a self-supervised manner on the outputs of the proposed SegFit.}
\label{tab:finetuning_improvements}
\end{table}

\subsection{Self-Supervised Fine-tuning}
Finally, we show that our refined body-part segmentations can be leveraged to automatically generate training data for segmentation models.
The original Human3D~\cite{Takmaz_2023_ICCV} model is fine-tuned using the real-world EgoBody~\cite{zhang2022egobody} dataset, which provides carefully curated human body-part annotations on 3D point clouds.
Our approach offers an automated method for generating such annotations.

As shown in Table~\ref{tab:finetuning_improvements}, fine-tuning Human3D with these additional labels further enhances its segmentation performance. We conduct experiments on both BEHAVE~\cite{bhatnagar22behave} and InterCap~\cite{huang2024intercap}, splitting each dataset into separate training and test sets.
Human3D is fine-tuned on the automatically generated body-part labels from the training set and evaluated on the test set.
To ensure high quality training data, we exclude scenes where human poses are poorly aligned, as determined by the error defined in Eq.~\ref{eq:obj} exceeding a predefined threshold.

Across both datasets, fine-tuning with our method significantly improves Human3D’s performance in terms of vertex-to-vertex error and segmentation accuracy. This highlights the effectiveness of our approach in refining part segmentation networks for new datasets.

\begin{table}[t]
    \centering
    \small
    \setlength{\tabcolsep}{2pt}
    \begin{tabular}{l cc cc}
    \toprule
                                  & V2V Orig. & V2V Filt. & Seg. Acc. Orig. & Seg. Acc. Filt. \\
    \textbf{Dataset} $\downarrow$ & [mm]      & [mm]      & [\%]            & [\%] \\
    \midrule
    \textbf{BEHAVE} & 37.4 & \textbf{30.1} & 90.44 & \textbf{92.35} \\
    \textbf{InterCap} & 142.0 & \textbf{127.3} & 68.35 & \textbf{74.33} \\
    \bottomrule
    \end{tabular}
    \caption{Mean vertex-to-vertex (V2V) error and segmentation accuracy of pseudo ground truths before and after excluding the 20\% of point clouds with the highest final loss in SegFit.}
    \label{tab:filtering}
\end{table}

\subsection{Analysis Experiments}
\label{sec:experiments_analysis}
Next, we conduct an ablation study to evaluate the impact of key components in our method, along with a hyperparameter analysis to examine the effect of loss weights.

\noindent\textbf{Ablation Study.}
To assess the contributions of individual components of our method, we conduct an ablation study and show the results of four different variants in Table~\ref{tab:segfit_analysis}, where \Circled{4} is the full \name{} model.

\Circled{1} A fundamental yet informative baseline for \name{} is an optimization process that does not incorporate body-part segmentation. This baseline serves to isolate and quantify the specific contribution of segmentation to the overall performance. Instead of leveraging body-part segments, the model is fitted using a more generic approach, where it is initialized with four different orientations to address potential symmetry ambiguities. When compared to the full method, this segmentation-free approach results in a notable decline in performance across all evaluated metrics and datasets. These findings highlight the crucial role of body-part segmentation in improving human pose and shape estimation, demonstrating its effectiveness in guiding the optimization process toward more accurate results.

\begin{figure*}
    \centering
    \includegraphics[width=0.99\linewidth]{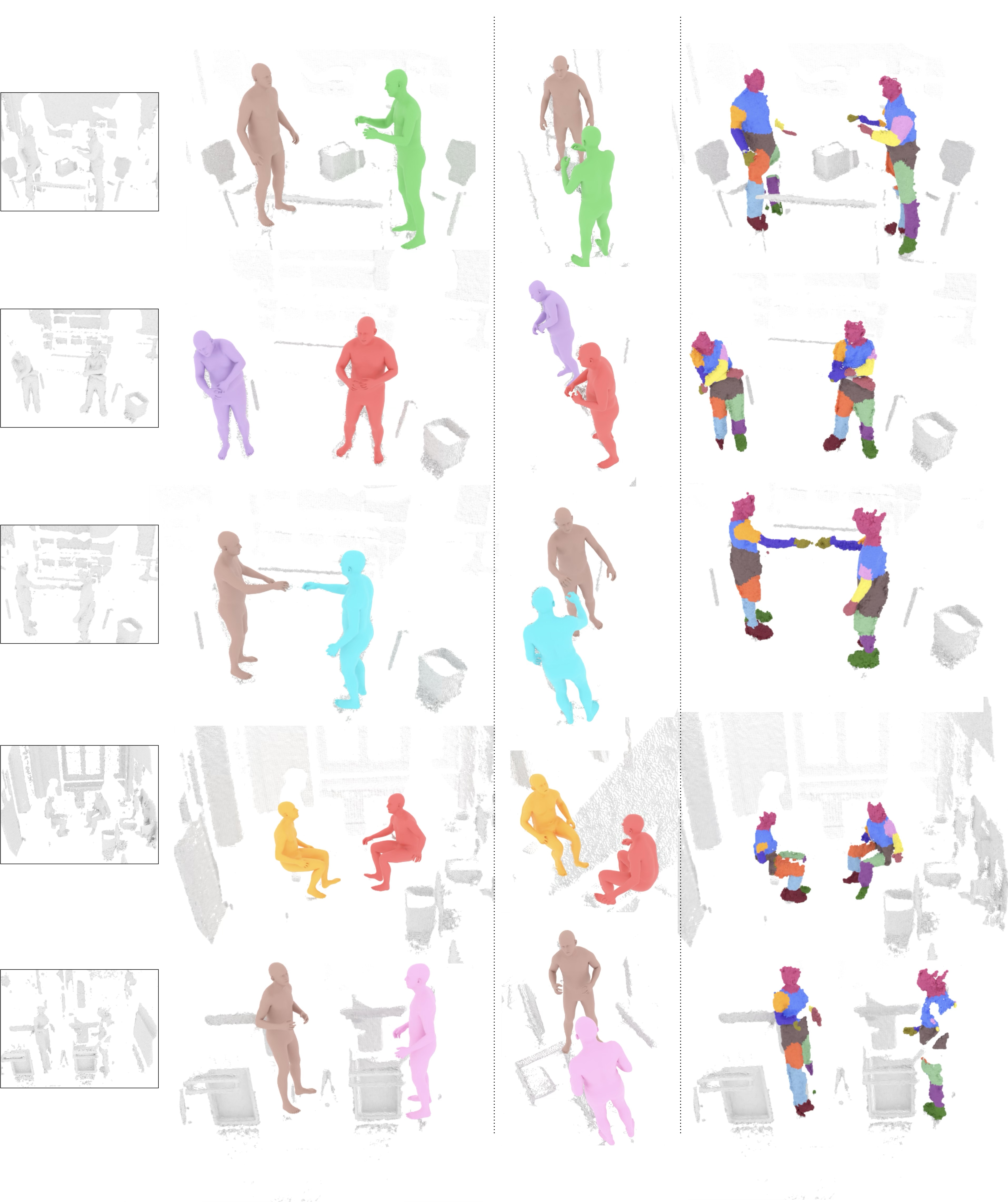}\\
        \vspace{-25px}
\begin{tabular}{cccc}
\vspace{-5mm}
 \textit{Input Scene} & \textit{SegFit Mesh (Ours)} & \textit{(Sideview)} & \textit{SegFit Body-Parts (Ours)}\\
 \hspace{2cm} &
 \hspace{5cm} &
 \hspace{3cm} &
 \hspace{6cm} \\
\end{tabular}
    \caption{\textbf{Qualitative Results.} Example outputs of our \name{} on the EgoBody~\cite{zhang2022egobody} dataset. From left to right: the input single-view point cloud showing the full scene including multiple humans, clutter and background, registered human meshes by \name{} from the front and side perspective, the refined body-part segmentation by \name{}. See Section \ref{sec:experiments_qualitative} for additional details.}
    \label{fig:quali1}
\end{figure*}

\Circled{2} The second baseline adds body-part segmentation during optimization while omitting the centroid-based initialization step. This modification leads to a significant reduction in errors compared to the first baseline, which does not utilize segmentation, and brings the accuracy closer to that of the full method across all datasets.
This variant also demonstrates a substantial boost in computational efficiency, reducing the runtime by at least a factor of two across all datasets.
However, we observed that a key limitation of this approach is its slower convergence, which primarily arises due to misalignment of body limbs. Specifically, points from the inner side of an arm (or leg) can sometimes be incorrectly matched to points on the outer side (or vice-versa), leading to incorrect correspondences that hinder optimization. These misalignments can prolong the fitting process and reduce overall robustness. This observation motivated our introduction of centroid-based initialization, which provides a more structured starting point for optimization, improving both alignment accuracy and convergence speed.

\Circled{3} 
Finally, we evaluate the accuracy of the centroid initialization step in isolation, without any subsequent optimization. While this approach results in an error approximately three times higher than the full method, it maintains a stable accuracy across all four datasets. Notably, it offers a significant speed advantage, with fitting times reduced by a factor of two to seven, averaging around half a second. This balance between speed and precision makes it a viable alternative for real-time applications where efficiency is prioritized over fine-grained accuracy.
\\

\begin{table}[t!]
\centering
\setlength{\tabcolsep}{2pt}
\begin{tabular}{ccccccc}
\toprule
 &\textbf{ B.P.} & \textbf{Cent.} & \textbf{Metric} & \textbf{BEHAVE} & \textbf{EgoBody} &  \textbf{InterCap}\\
\midrule
\Circled{1} & \xmark & \xmark & & $55.6$ & $109.7$ & $206.8$ \\
\Circled{2} & \cmark & \xmark & V2V & $39.9$ & $48.5$ & $147.8$ \\
\Circled{3} & \xmark & \cmark & {in mm} & $97.3$ & $105.2$ & $154.7$ \\
\Circled{4} & \cmark & \cmark & & $\mathbf{37.0}$ & $\mathbf{47.9}$ & $\mathbf{147.2}$ \\
\midrule
\Circled{1} & \xmark & \xmark & & $43.6$ & $100.3$ & $195.4$ \\
\Circled{2} & \cmark & \xmark & MPJPE & $42.6$ & $42.7$ & $141.3$ \\
\Circled{3} & \xmark & \cmark & {in mm} & $80.8$ & $87.9$ & $148.5$ \\
\Circled{4} & \cmark & \cmark & & $\mathbf{42.2}$ & $\mathbf{20.6}$ & $\mathbf{140.7}$ \\
\midrule
\Circled{1} & \xmark & \xmark & & $6.05$ & $4.28$ & $10.7$ \\
\Circled{2} & \cmark & \xmark & Time & $2.98$ & $3.70$ & $6.21$ \\
\Circled{3} & \xmark & \cmark & { in s} & $\mathbf{0.40}$ & $\mathbf{0.40}$ & $\mathbf{0.66}$ \\
\Circled{4} & \cmark & \cmark & & $1.86$ & $1.79$ & $1.37$ \\
\bottomrule
\end{tabular}
\caption{Ablation study analysing the effect of body parts (B.P.) and Centroids (Cent.) on the BEHAVE~\cite{bhatnagar22behave}, EgoBody~\cite{zhang2022egobody}, and InterCap~\cite{huang2024intercap} datasets. Metrics are vertex-to-vertex (V2V) distance, mean-per-joint-position-error (MPJPE) and runtime.}
\label{tab:segfit_analysis}
\end{table}

\paragraph{Hyper-parameter Analysis.}
Table~\ref{tab:regularization_weights} examines the impact of the weighting coefficients $\lambda_{\text{pose}}$ and $\lambda_{\text{shape}}$ on the pose and shape terms. We conduct a grid search over the values ${0.0, 0.5, 1.0, 2.0}$ and find that the lowest V2V error is achieved when both coefficients are set to $0.5$. Notably, setting $\lambda_{\text{shape}} = 0$ underscores the importance of the shape term, which prevents unnatural body deformations.

\newcommand{\colcell}[1]{
  \pgfmathsetmacro{\normalized}{(#1 - 41.8) / (52.2 - 41.8)}
  \pgfmathsetmacro{\green}{(1 - \normalized) * 100}
  \pgfmathsetmacro{\red}{\normalized * 100}
  \xdef\colorval{lightgreen!\green!lightred}
  \cellcolor{\colorval}{$#1$}
}   

\definecolor{lightgreen}{HTML}{90EE90}
\definecolor{lightred}{RGB}{255, 182, 193}

\begin{table}[t!]
    \centering
    \begin{tabular}{c cccc}
        \toprule
        $\lambda_{\text{shape}} \backslash \hspace{1px} \lambda_{\text{pose}}$
        & $0.0$ & $\mathbf{0.5}$ & $1.0$ & $2.0$ \\
        \midrule
        $0.0$ & \colcell{52.2} & \colcell{45.0} & \colcell{47.6} & \colcell{48.7} \\
        $\mathbf{0.5}$ & \colcell{51.6} & \cellcolor{lightgreen}$\mathbf{41.8}$ & \colcell{42.5} & \colcell{44.1} \\
        $1.0$ & \colcell{51.6} & \colcell{42.9} & \colcell{44.3} & \colcell{45.2} \\
        $2.0$ & \colcell{51.8} & \colcell{44.1} & \colcell{45.7} & \colcell{46.9} \\
        \bottomrule
    \end{tabular}
    \caption{Effect of $\lambda_{\text{shape}}$ and $\lambda_{\text{pose}}$ on the vertex-to-vertex (V2V) error (in mm). Best value for both hyper-parameters is $0.5$.}
    \label{tab:regularization_weights}
\end{table}

\subsection{Qualitative Results and Discussion}
\label{sec:experiments_qualitative}

Figure~\ref{fig:quali1} presents several representative examples of \name{} applied to the EgoBody~\cite{zhang2022egobody} dataset. The input scenes and corresponding human poses exhibit significant diversity, introducing multiple challenges such as occlusions caused by scene clutter, partial visibility due to single-view depth sensors, and artifacts from the scanning process. In many cases, even for a human observer, it is difficult to discern which points belong to a person or to specific body parts based solely on the raw input data.
Despite these challenges, our method demonstrates robust performance, successfully recovering human poses even under severe occlusions. The bottom example highlights this capability, emphasizing the importance of leveraging body-part segmentation to handle partial and noisy real-world point clouds. However, our approach is not without limitations. One common failure mode occurs when human instances are entirely missed during segmentation, resulting in missing reconstructions. Another frequent issue is inaccurate limb registration, particularly in cases where subjects cross their arms or legs, as seen in the second example from the top. These ambiguities can lead to incorrect alignments, particularly in highly occluded settings.
To further enhance pose estimation in challenging scenarios, integrating additional scene reasoning, particularly with affordance-based constraints \cite{delitzas2024scenefun3d}, could help refine predictions when subjects closely interact with their environment, where occlusions are most severe.
\section{Conclusion}

We introduce SegFit, a novel hybrid approach for fitting parametric human body models to diverse 3D point clouds,
combining body part segmentation and human pose and shape priors to iteratively enhance both segmentation and pose fitting accuracy.
Future work will explore potential improvements to the pose fitting accuracy, such as by introducing a penetration loss term for scenes where humans interact with each other or with objects in their environment.

\paragraph{Acknowledgments.}
This project is partially supported by an SNSF PostDoc.Mobility fellowship.

{
    \small
    \bibliographystyle{ieeenat_fullname}
    \bibliography{main}
}

\end{document}